\documentclass[]{svmult}

\usepackage{mathptmx}       
\usepackage{helvet}         
\usepackage{courier}        
\usepackage{type1cm}        
\usepackage{makeidx}         
\usepackage{graphicx}        
\usepackage{multicol}        
\usepackage[bottom]{footmisc}

\makeindex             

\usepackage{hyperref}
\usepackage{cite}
\usepackage{multirow}

\begin{document}

\title*{Labelling Drifts in a Fault Detection System for Wind Turbine Maintenance}

\author{I\~nigo Martinez, Elisabeth Viles, I\~naki Cabrejas}
\institute{I\~nigo Martinez · I\~naki Cabrejas \at NEM Solutions, 20009 San Sebastian, Spain \newline \email{imartinezl@alumni.tecnun.es} \newline 
\and Elisabeth Viles \at University of Navarra - Tecnun, 20018 San Sebastian, Spain \newline \email{eviles@tecnun.es}   \newline
}

\maketitle

\abstract{
A failure detection system is the first step towards predictive maintenance strategies. A popular data-driven method to detect incipient failures and anomalies is the training of normal behaviour models by applying a machine learning technique like feed-forward neural networks (FFNN) or extreme learning machines (ELM). However, the performance of any of these modelling techniques can be deteriorated by the unexpected rise of non-stationarities in the dynamic environment in which industrial assets operate. This unpredictable statistical change in the measured variable is known as concept drift. In this article a wind turbine maintenance case is presented, where non-stationarities of various kinds can happen unexpectedly. Such concept drift events are desired to be detected by means of statistical detectors and window-based approaches. However, in real complex systems, concept drifts are not as clear and evident as in artificially generated datasets. In order to evaluate the effectiveness of current drift detectors and also to design an appropriate novel technique for this specific industrial application, it is essential to dispose beforehand of a characterization of the existent drifts. Under the lack of information in this regard, a methodology for labelling concept drift events in the lifetime of wind turbines is proposed. This methodology will facilitate the creation of a drift database that will serve both as a training ground for concept drift detectors and as a valuable information to enhance the knowledge about maintenance of complex systems.
}


\section{Introduction}
\label{sec:1}

The prevailing competitive marketplace demands companies from asset-intensive industries to create cost-efficient processes, both in manufacturing and in maintenance. Unforeseen breakdowns not only can cause expensive downtime, but also safety and environmental detriments that may lead to injuries or fatalities, as well as enormous legal expenses. In this sense, the ability to forecast machinery failure is vital for reducing maintenance costs, operation downtime and safety hazards. It is therefore essential to develop failure detection techniques to monitor the health of a system, which are encompassed in Condition Based Maintenance (CBM) strategies \cite{Jardine2006, Heng2009} and more recently, in prognostics and health management (PHM) \cite{Zio2012,Vichare2006, Cheng2010}

A failure management system consists of several interrelated modules. \cite{Salfner2010} At first, failure prone situations are identified, using a continuous measure that judges the current situation as more or less failure prone; this is known as the failure detection stage. Then, after failure detection, the diagnosis stage is invoked, in order to find out where the error is located and where its root cause may be. Once the diagnosis is completed, maintenance actions are scheduled.

This article focuses in the failure detection module, which provides real-time monitoring of industrial processes by forecasting machinery health based on condition data and predicting possible incipient failures. Several industrial sectors have adopted these type of systems to improve their maintenance processes and to manage their assets more effectively. In particular, we are interested on applications to the maintenance of wind turbines, where various strategies have already been developed. \cite{Yang2013, ShuangwenSheng2011}

Technical approaches for building models in incipient failure detection systems can be categorized broadly into data-driven, model-based, and hybrid approaches. More details regarding the mentioned techniques can be found at \cite{Al-Turki2014}. This study deals with a pure data-driven approach, where information coming in real-time from different sensors is taken into account, and detection of possible anomalies is provided after learning the normal and expected behaviour from the industrial assets. 

By using data sources, different strategies can be followed to build normal behaviour models, such as stochastic models, machine learning algorithms, Bayesian and fuzzy classifiers, time series prediction or pattern recognition. In the analyzed failure detection system a state-of-the-art neural network, Extreme Learning Machine (ELM), has been applied, due to its ability to easily model dynamic non-linear behaviours \cite{Kubat1999}, and also because of its wide use in the prognostics of industrial systems and wind turbines \cite{Liu2012, Pelletier2016, Qian2015, Qian2017, Saavedra-Moreno2013, Wan2014}

Whatever the selected model may be, an initial training batch is necessary for learning relationships between variables. Taking a feed-forward neural network as example, the network is trained with an historical data set, a fixed and static information about the past events of the system. By comparing the expected behaviour with its real and current functioning, both a normal behaviour deviation degree as well as an estimation certainty degree are obtained. These are used to recognize an anomaly, which afterwards can be related to a known failure mode.\cite{Garcia2006} These failure detection strategies are based on the hypothesis that any asset should behave similarly under resembling conditions, concluding that a deviation from the model should speak for a symptom of dysfunction.

However, the performance of any of these modelling techniques can be deteriorated by the unexpected rise of non-stationarities in the dynamic environment in which industrial assets operate. Unpredictable statistical changes in the measured variable are desired to be detected by means of statistical detectors and window-based approaches. In order to evaluate the effectiveness of current detectors and also to design an appropriate novel technique for an specific wind turbine maintenance application, it is essential to dispose beforehand of a characterization of the existent drifts. Under the lack of information in this regard, in this article a methodology for labelling concept drift events in the lifetime of wind turbines is proposed.

Having introduced the necessity for predictive maintenance and failure detection systems on Sect. \ref{sec:1}, the rest of the article is structured as follows: the irruption of non-stationary data in failure detection systems is explained on Sect. \ref{sec:2}, along with the main adaptation techniques to this issue. On Sect. \ref{sec:3}, an expert-based platform for drift labelling is described. Finally, Sect. \ref{sec:4} presents the main conclusions and paths for future work and improvement.

\section{Non-stationary Data in Traditional Machine Learning}
\label{sec:2}
One of the problems that arise from the presented failure detection systems is that data is expected to be independent and identically distributed (i.i.d) along all the lifetime of the asset \cite{Gama2014}. Unfortunately, in real industry applications, this assumption does not often hold true. In fact, due to the non stationarity in the monitored variables of the assets, the normality models need to be checked periodically and actions need to be taken to adapt to the new normality \cite{Zliobaite2010}.

Regarding failure detection systems, the non stationarity can be due to multiple causes, such as sensors recalibration, replacement of a component during a corrective maintenance intervention, the wearing of some mechanical component, etc. The causes of the non-stationary data are critical to decide whether to take actions or not. Yet, model adaptation should only occur if the detection quality is seriously affected. An illustrative example can help understand this issue: the wearing of components or the lost of efficiency represent events that are desired to detect by studying deviations from the normality model, implying that the model should not be updated. 
Unfortunately, the representation of these events vary enormously from asset to asset and from time to time. This high variability implies some difficulties to find patterns that help to differentiate between a need-to-update situation and a maintain-model one.

Even though failure prediction systems are very vulnerable to non-stationarities, at the moment this issue being solved with human supervision. However, companies are monitoring more and more assets each year, and they are starting to reach up to a point where a manual checking is unfeasible. As an example, the recreation of the normal wind turbine operation can consist of about 10 models of subsystems or critical parts of the wind turbine. For 1.000 wind turbines, an architecture of 10,000 normal behaviour models would be managed. Moreover, taking into account current trends, with more wind farms being installed worldwide each year, it is necessary to include an automated system that guarantees the normality of each model for the current situation. This automated system would reduce as much as possible the human intervention in the system, thus allowing the scalability of the detection technology.

\subsection{Concept Drift Theory}
\label{sec:2;subsec:1}
Current artificial intelligence and machine learning based applications live under the assumption that the systems they predict are static and stationary \cite{Gama2014}, neglecting that in reality, the world and the data generating processes are often dynamic and non-stationary. Unexpected changes in the environment or perturbed, incorrect and missing data can statistically detach the measured variable from the monitored feature, inducing to wrong decisions. Moreover, the spreading of online deployments with learned models gives increasing urgency to the development of efficient and effective mechanisms to address learning in the context of non-stationary distributions, or as it is commonly called concept drift \cite{Zliobaite2010, Webb2016, Tsymbal2004, Hoens2012, Gama2014, Ditzler2015}. Hence, concept drift can be defined as the unpredictable statistical change in measured variables, making the static and time-invariant model useless. 

A concept is statistically defined on the literature \cite{Gama2014} as the joint probability distribution of predictors (independent) variables and response (dependent) variables at a given period $t$: $\,Concept = P_{t}(X,Y)\,$, where $X$ denotes a random variable over vectors of predictor values, and $Y$ represents a random variable over the output or response.  A concept drift occurs whenever a pair of periods of time $t$ and $u$ depict different joint distributions: $P_{t}(X,Y) \neq P_{u}(X,Y) $. According to Bayesian theory
, a change in the joint probability distribution can occur due to a change either in the prior probability distribution $P_{t}(Y)$, in the class conditional probability distribution $P_{t}(X|Y)$, or in the posterior probability distribution $P_{t}(Y|X)$. 

There is the implicit assumption that drift occurs over discrete periods of time, bounded before and after by stable periods without drift \cite{Webb2016}. Under this assumption, the speed of drift can be quantified \cite{Gama2014} as sudden if there is a sharp boundary, as gradual if the transition is smooth, and reoccurring if drifts repeat over time.

A quantitative characterization of drifts can be found in\cite{Webb2016}. These measurements would help to create a taxonomy of drifts for a given application. Here the most relevant measures of drift are summarized:
\begin{description}[Magnitude 1]
\item[\textbf{Magnitude}]{Distance between the concepts at the start $t$ and end $u$ of the period of drift. The metric used is usually the Hellinger Distance or the Total Variation Distance.  
\begin{equation}
Magnitude_{t,u}=D(t,u) 
\end{equation}
For example, the Hellinger Distance $H$ is a metric that measures the difference between two probability distributions $P$ and $Q$, and can be defined as:
\begin{equation}
H_{P,Q}=\frac{1}{\sqrt{2}}\,||P-Q||^2 
\end{equation}
}
\item[\textbf{Duration}]{The elapsed time over which a period of drift occurs. This measurement is critical for the differentiation of a fault an a concept drift. \begin{equation}
Duration_{t,u}=u-t
\end{equation}}
\item[\textbf{Path Length}]{Length of the path that the drift traverses during a period of drift, i.e. the sum of the drift magnitude between pairs of consecutive points.
\begin{equation}
PathLen_{t,u}=\lim_{n\to\infty} \sum_{k=0}^{n-1}D\Big(t+\frac{k}{n}(u-t),t+\frac{k+1}{n}(u-t) \Big)
\end{equation}}
\end{description}

\subsection{Adaptation Techniques}
\label{sec:2;subsec:2}
Under the presence of concept drift, passive or active solutions are available in order to prevent a loss of performance in the detection quality or in the prediction accuracy. On one hand, passive strategies continuously adapt the model without the need of detecting a change.  They aspire to maintain an up-to-date model at all times, either by retraining the model on the most recently observed samples, or by enforcing an ensemble of classifiers \cite{Krawczyk2017}. As the stability-plasticity dilemma defines, a single-classifier model is not able to retain existing knowledge and at the same time learn new information, so ensembles of classifiers are usually put to work.\cite{Mouret2015} Ensembles continuously update the weights of the fusion rule or create and remove models from the pool, being more accurate than single-classifier models, and easily incorporating new data or forgetting irrelevant knowledge. 

However, in a failure detection system, the prediction accuracy is expected to decrease when a failure appears, and thus the objective is not to improve the performance, but to preserve the detection quality. In addition, passive adaptation techniques allow the normality model to learn from anomalies and failures as well, thus expanding the decision boundary and reducing the detectability of the system. Consequently, passive solutions should be discarded for failure detection applications, and active solutions should be preferable. 

Active solutions rely on triggering mechanisms, that is, detection methods that indicate whether a drift has occurred or not based on a change in the statistics of the data-generating process. Change-detection tests \cite{Goncalves2014, Sobolewski2013, Sebastiao2009, Santos2016, Pears2014, Ross2012} can be triggered either by monitoring the distribution of unlabeled observations or by a change in the prediction performance of the model. Whatever the monitored signal may be, these methods explicitly localize the change point in time, and invoke the substitution of the model with a new one, trained with recent data, that maintains the prediction accuracy and the overall detectability. In this article the attention is onto the active detection of concept drifts, whereas posterior actions, such as the retraining of the model are not studied.

\section{Proposed Methodology: Drift labelling}
\label{sec:3}
Failure detection technologies that take into consideration deviations from normal behaviour models will suffer from the concept drift problem as time passes on. This maintainability problem of failure detection technologies is in fact a reality in industrial applications. Appropriate adaptation techniques to tackle concept drifts can only be designed if there is a previous knowledge of the types of drifts that can appear. Therefore, this research rises from the current lack of knowledge regarding drift detection in real industrial assets, where a taxonomy of drifts is hard to discern and theoretical drift detectors have unknown performances.

In order to address the problem of normality changes in the performance of complex systems this article proposes a methodology for the identification, record keeping, and posterior classification of drifts in the specific case study of wind turbines maintenance. The result is the development of a platform that allows experts to create a reference dataset of behaviour changes in the normal performance of wind turbines. This will serve as training ground for the posterior classification of such changes, and also to test the effectiveness of those detectors in drift situations. At the same time, this will enhance the knowledge about the operation of the wind turbines, and as a consequence, the improvement of the applied failure detection systems.

\subsection{Platform Overview}
\label{sec:3;subsec:1}
\begin{figure}[!hbtp]
\centering
\includegraphics[width=0.8\linewidth]{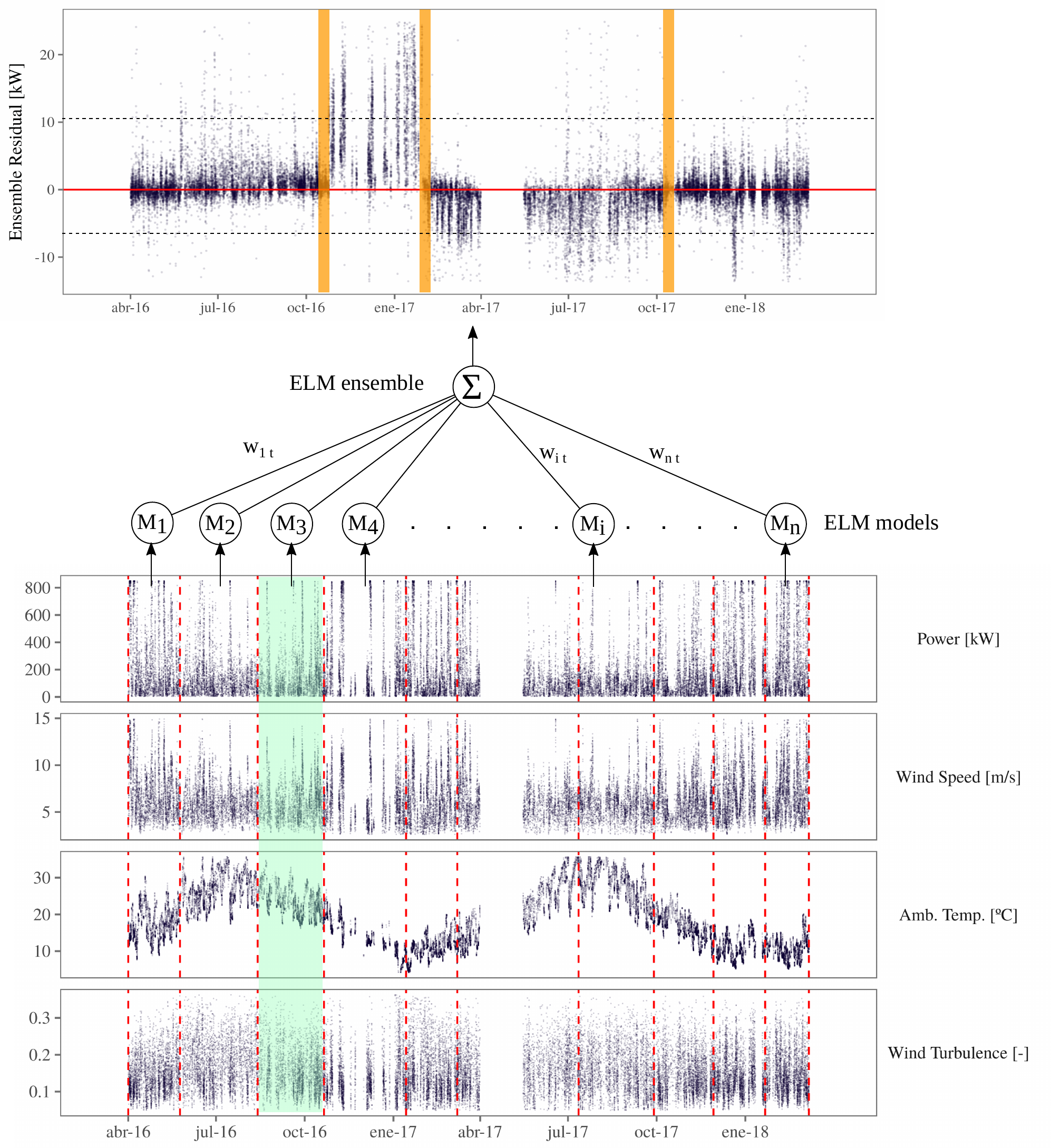}
\caption{Each ELM model is trained with independent batches and combined onto the ensemble}
\label{fig:1}
\end{figure}

The process starts by collecting data from the wind turbines. In this regard, the operating and environmental conditions of virtually all wind turbines in operation today are recorded by the turbines' supervisory control and data acquisition (SCADA) system in 10-minute intervals \cite{Bangalore2018}. The number of signals available to the turbine operator varies considerably between different manufacturers as well as between generations of turbines by the same manufacturer. The recreation of the normal wind turbine operation can consist of 10 models of critical parts, each of them monitoring a known variable. In this article the wind turbine's power model is illustrated as example (Fig. \ref{fig:1}), where the ambient temperature, the wind speed and the wind turbulence act as predictors. These variables are available in almost all SCADA systems.

\begin{figure}[!hbtp]
\centering
\includegraphics[width=0.7\linewidth]{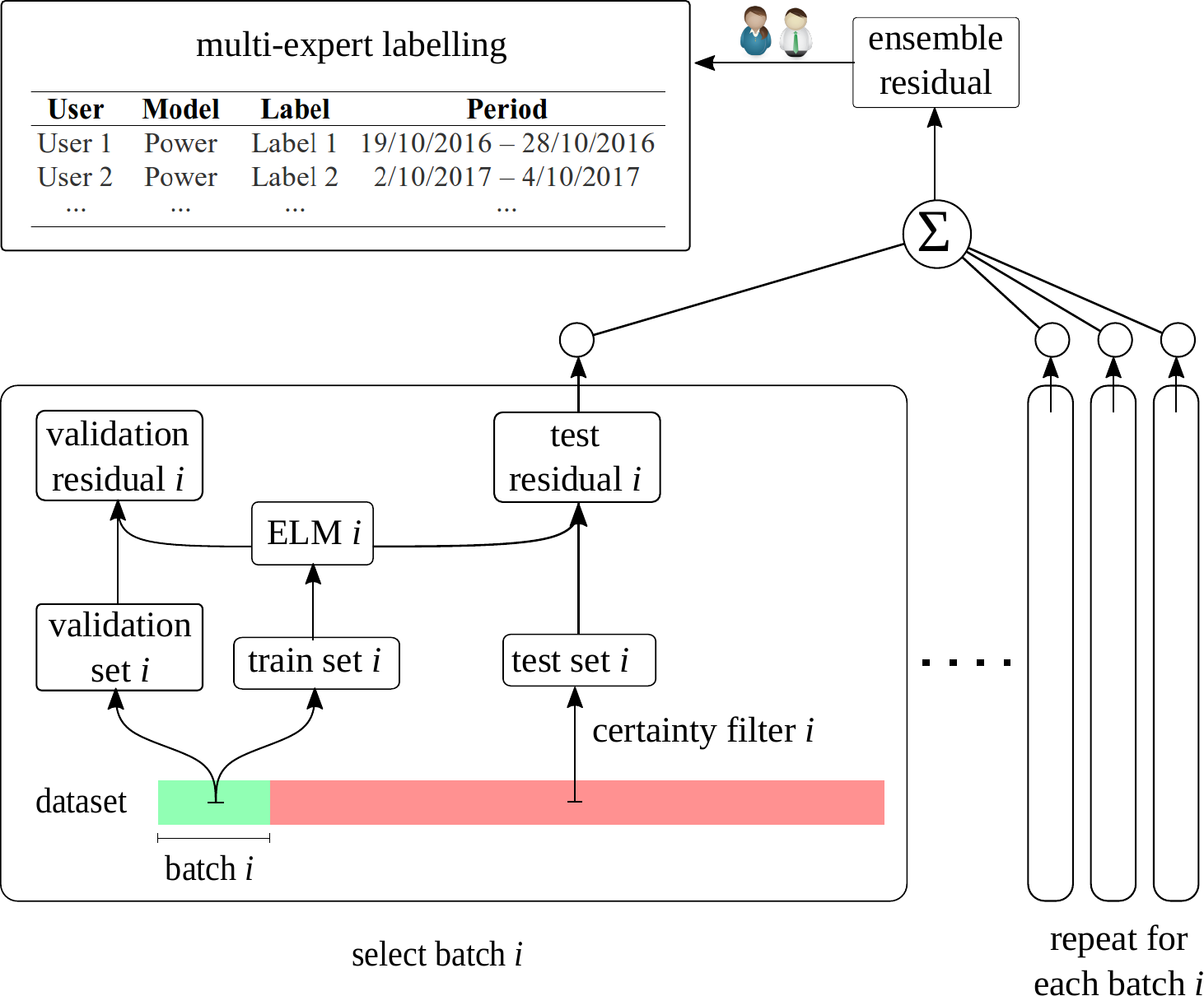}
\caption{Overview of the multi-expert platform. Experts label periods of drift on the ensemble residual. Each ELM model contributes to the ensemble, and is trained with an independent batch.}
\label{fig:2}
\end{figure}

As was stated on Sect. \ref{sec:2;subsec:1}, drifts can take diverse forms (sudden, gradual...) and appear on any variable (inputs, output). The proposed method tries to identify changes in the posterior probability distribution $P_{t}(Y|X)$, that is, in the output-inputs relationship, since those changes directly affect the performance of the failure detector. Therefore, the normal behaviour model residual ---difference between the actual power and the predicted one--- will be shown to the expert.

In order to automatize the labelling, an interactive web platform has been developed using the Shiny package in R. This platform shows the model residual as a time series, and experts are allowed to label any drift or event by dragging the desired period on the chart. The selected period is then classified, based on the severity of the drift and the possible causes (sensor mis-calibration, maintenance action, power limitation, etc.). Each time a period is labeled, it will be visualized on the chart and registered on the database, along with the expert-user and model information.

Due to the fact that this platform uses visual inspection to label periods of drift, the results can be subjected to the expert's own criteria. In subsequent work, expert's opinion will be complemented with information coming from the asset's maintainers, such as maintenance work orders. In order to evaluate appropriately the effectiveness of the labelling system and reduce this source of uncertainty, redundancy between multiple experts is proposed. Apart from allowing multiple users to label the same cases, each user will indicate a qualitative measure of confidence about their labels.  

\subsection{ELM Ensemble}
\label{sec:3;subsec:2}
Normal behaviour models are created by training a single layer extreme learning machine (ELM), which has the same topology as a single layer FFNN, with the difference that the hidden layer weights are not tuned, but randomly assigned. This slight modification makes ELM training extremely fast, since only output weights need to be optimized, which is worked out by a simple ridge regression \cite{Huang2006}. ELM have already been used to model the real operation of wind turbines, showing great generalization properties \cite{Wan2014, Saavedra-Moreno2013, Qian2015, Qian2017}

In order to make the model predictions more robust and trustworthy, an ensemble of models \cite{Lan2009} is used rather than an individual model. Each model from the ensemble is trained with a different batch of data, which will avoid two issues:
\begin{itemize}
\item On one side, the training data can contain failures and behaviour changes, which should not be learned. Rather than selecting a fault-free training set, the generalization capability of the weighted combination of individual models will automatically not learn anomalies and features that rarely appear, but the general behaviour, also guaranteeing a low validation error. 

\item On the other side, when a subset of the whole data is used as training set, it is possible that it may not completely represent the entire numerical space of the variables. A model should only predict the response when there is enough certainty about the inputs, otherwise the estimation may not be reliable. An example of such situation can be seen with the ambient temperature, where a clear trend occurs along the year. The entire year should be selected in order to cover all temperature values and make certain predictions. 
\end{itemize}

The residual calculation process has been illustrated on Fig. \ref{fig:1} . Each individual ELM model is trained with a fixed size batch. A random sample of the batch is used for validation, whereas the remaining trains the model. The rest of the dataset is first filtered by certainty to the training set, and then is passed to the model to predict the response. The certainty filter makes sure the inputs are known to the model. 
This process is repeated for each batch, as has been represented on Fig. \ref{fig:2},  thus obtaining the ensemble of models, and the combination of residuals, which is shown to the experts.

\subsection{Preliminary Results}
\label{sec:3;subsec:3}

A comparison of the labelled periods of drift with the triggering of some active detectors has been included in Table \ref{table_1}. A group of 4 experts labelled events from 98 wind turbines ranging 3 years of data. Overall, the experiment showed that even though detectors match the manually labelled periods, captured by a high sensitivity, low precision results put into evidence that detectors trigger when other type of drifts happen or when a noisy signal undermines the performance of the detectors.

\begin{table}[]
\centering
\resizebox{\textwidth}{!}{%
\begin{tabular}{lcccccccccc}

\hline
\multicolumn{1}{l}{} & ADWIN & CUSUM & GMA & HDDM\_A & HDDM\_W & PH & SEED & SeqDrift1 & SeqDrift2 & STEPD \\
\hline
Precision & 0.422 & 0.408 & \textbf{0.571} & 0.304 & 0.432 & 0.413 & 0.370 & 0.412 & 0.519 & 0.261 \\
Sensitivity & 0.711 & \textbf{0.816} & 0.316 & 0.737 & 0.500 & 0.684 & 0.789 & 0.737 & 0.737 & \textbf{0.816} \\

\hline
\end{tabular}
}

\caption{Precision and sensitivity of other detectors on manually labelled periods, which are assumed to be the true condition.The bold numbers indicate the maximum of each metric.}
\label{table_1}
\end{table}
The detectors were implemented using MOA, an open-source framework \cite{DBLP:journals/jmlr/BifetHKP10} for dealing with massive evolving data streams.
The optimal parameters for each detector were extracted from the article by Gon\c{c}alves et. al \cite{Goncalves2014}. The definition for the tested detectors can be found in \cite{Gama2014, Goncalves2014,Sobolewski2013, Sebastiao2009, Santos2016, Pears2014, Ross2012}.

\section{Conclusions and Further Work}
\label{sec:4}

In this article, a methodology for the identification, record keeping, and posterior classification of drifts in a wind turbine maintenance case has been presented. In real complex systems, normality changes are not as clear as in artificial datasets, so it is difficult to implement active adaptation techniques, such as drift detectors. By following the proposed method, a database of behaviour changes can be created with the help of experts, serving both as a training ground for concept drift detectors and as a valuable information to enhance the knowledge of complex systems.

In addition, with this novel method, different types of drifts -- sudden, gradual, recurrent -- can be classified with a certain degree of objectivity. From the obtained labels, drift features such as magnitude, duration and path length can be estimated as well.

As was stated on Sect. \ref{sec:1}, the problem of concept drift can be found in any failure prediction system that uses deviations from a normal behaviour model to identify incipient failures. The methodology explained on Sect. \ref{sec:3} has only been applied to a wind farm maintenance case. This methodology should be easily extended to other complex systems where no information about the possible behaviour changes exists. 

After obtaining the characterization of drifts, the effectiveness of current drift detectors has been evaluated for this specific industrial application. A future line of research will focus on the design of novel drift detection techniques, to be tested in real-world environments with the purpose of improving the quality of the current drift detectors, tested in Section \ref{sec:3;subsec:3}. In this way, an ensemble of detectors \cite{Maciel2016, Sobolewski2013b, Wozniak2017, Du2015} could provide robustness in real life applications, where the representation of drift events vary enormously from asset to asset and from time to time. Just like individual classifiers are limited by the stability-plasticity dilemma, a single detector is not able to discern several types of drifts. In fact, most concept drift detectors found on the literature  are based on assumptions about the distribution of the variables and try to identify changes whenever their stated hypothesis do not hold true.

\begin{acknowledgement}
This research has been supported by NEM Solutions, a technology-based company focused that provides intelligent maintenance of complex systems to O\&M businesses.
\end{acknowledgement}

\bibliographystyle{spphys}
\bibliography{bibtex}

\end{document}